\documentclass[a4, 10 pt, conference]{ieeeconf}  % Comment this line out if you need a4paper

\IEEEoverridecommandlockouts                              % This command is only needed if 
\usepackage{float}
\usepackage{graphicx}

\title{\LARGE \bf
Enhancing Consistency in Multimodal Dialogue System Using LLM with Dialogue Scenario
}

\author{Hiroki Onozeki$^{1}$, Zhiyang Qi$^{1}$, Kazuma Akiyama$^{1}$,\\Ryutaro Asahara$^{1}$, Takumasa Kaneko$^{1}$, and Michimasa Inaba$^{1}$% <-this % stops a space
\thanks{$^{1}$Graduate School of Informatics and Engineering, The University of Electro-Communications, Japan. Representative email address:
        {\tt\small o2330023@edu.cc.uec.ac.jp}}%
}

\makeatletter
\def\endthebibliography{%
        \def\@noitemerr{\@latex@warning{Empty `thebibliography' environment}}%
        \endlist
}
\makeatother

\begin{document}

\maketitle
\thispagestyle{empty}
\pagestyle{empty}

%%%%%%%%%%%%%%%%%%%%%%%%%%%%%%%%%%%%%%%%%%%%%%%%%%%%%%%%%%%%%%%%%%%%%%%%%%%%%%%%
\begin{abstract}

This paper describes our dialogue system submitted to Dialogue Robot Competition 2023.
The system's task is to help a user at a travel agency decide on a plan for visiting two sightseeing spots in Kyoto City that satisfy the user.
Our dialogue system is flexible and stable and responds to user requirements by controlling dialogue flow according to dialogue scenarios.
We also improved user satisfaction by introducing motion and speech control based on system utterances and user situations.
In the preliminary round, our system was ranked fifth in the impression evaluation and sixth in the plan evaluation among all 12 teams.

\end{abstract}

%%%%%%%%%%%%%%%%%%%%%%%%%%%%%%%%%%%%%%%%%%%%%%%%%%%%%%%%%%%%%%%%%%%%%%%%%%%%%%%%
\section{INTRODUCTION}

This paper describes our dialogue system submitted to Dialogue Robot Competition 2023 (DRC2023) \cite{overview}\cite{journals}.
In this competition, a humanoid robot is used to build a dialogue system to play the role of a counter salesperson in a travel agency.
The task of the dialogue system is to consult with the user and help the user decide on a tourism plan to visit two sightseeing spots in Kyoto City.
In this task, the system asks the user about his or her requirements regarding sightseeing spots, searches for sightseeing spots that fulfill the user's requirements, and recommends the sightseeing spots.
Therefore, a necessary step is to input the dialogue context and the information about sightseeing spots into the response generation model and generate appropriate responses that consider this information.
However, if the model receives a long input, the quality and consistency of the generated responses may be compromised because the model must understand many contexts at once when processing the information.

Therefore, we created a dialogue scenario to control the dialogue flow and generate consistent responses by inputting only the necessary dialogue context and sightseeing spot information into the model according to the dialogue scenario.
We also improved user satisfaction by introducing motion and speech control to emphasize important words based on system utterances and user situations.

%%%%%%%%%%%%%%%%%%%%%%%%%%%%%%%%%%%%%%%%%%%%%%%%%%%%%%%%%%%%%%%%%%%%%%%%%%%%%%%%

\section{SYSTEM ARCHITECTURE}

This section describes the main components of our system.

\subsection{Dialogue Scenario}

\begin{figure}[t]
	\begin{center}
		\includegraphics[width=\linewidth]{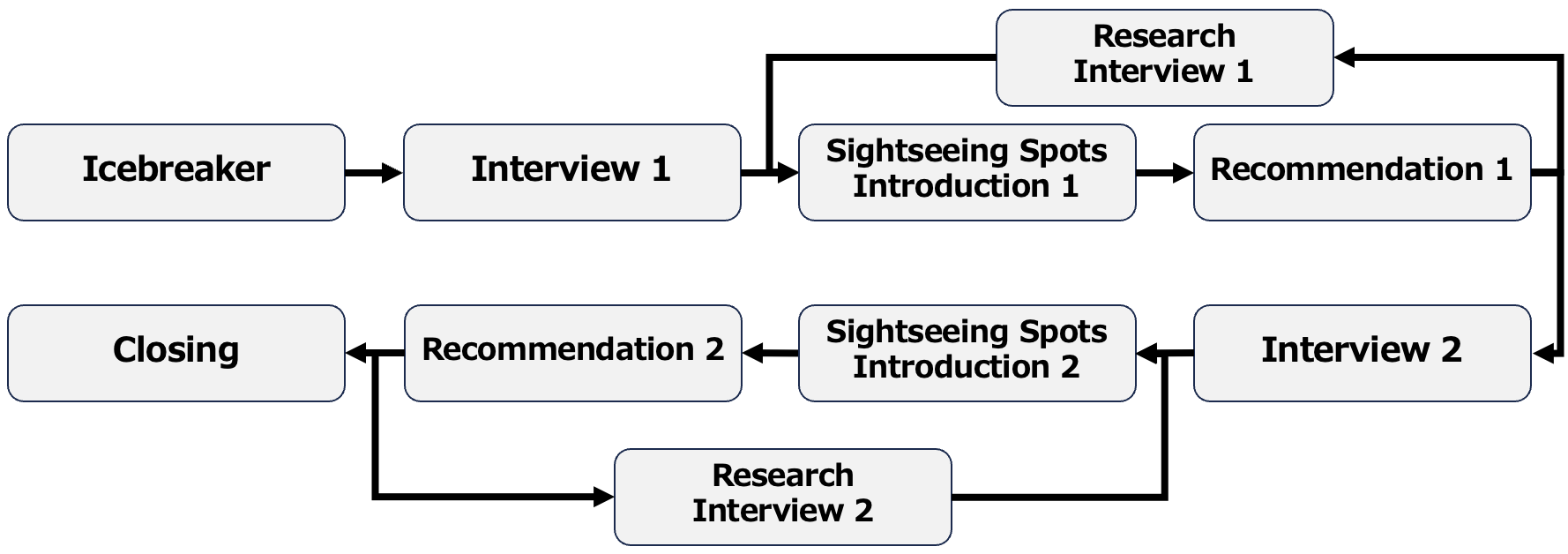}
		\caption{Dialogue Scenario}
		\label{fig:Dialogue_Scenario}
	\end{center}
\end{figure}

The dialogue scenario of our system is shown in Fig. \ref{fig:Dialogue_Scenario}.
In this dialogue scenario, the dialogue starts with a humorous chat as an Icebreaker to attract the user's interest.
Subsequently, the user's requirements regarding the first sightseeing spot are elicited in Interview 1.
The user's requirements are generated as keywords from the dialogue context of the interview using GPT-3.5-turbo and these keywords are entered into the sightseeing spot search API to search for three sightseeing spots.
The sightseeing spots are then presented with their image and the reason for the recommendation is briefly presented in Sightseeing Spots Introduction 1.
Next, Sightseeing Spots Recommendation 1 uses the information from the searched sightseeing spots to make recommendations and determine the first sightseeing spot the user should visit.
If none of the three suggested sightseeing spots is of interest to the user, the user's preferences for attractions are elicited again in Research Interview 1, and the sightseeing spot search is conducted again.
Once the first sightseeing spot to be visited by the user is determined, Interview 2, Sightseeing Spots Introduction 2, Recommendation 2, and Research Interview 2 are also conducted for the second sightseeing spot as well, and the second sightseeing spot to be visited by the user is determined.
Finally, the user confirms the sightseeing plan decided in Closing, and the dialogue ends.
By controlling the dialogue flow according to this dialogue scenario, the task of deciding the tourism plan can be performed stably without any dialogue breakdowns.

\subsection{Response Generation}

\begin{figure}[t]
	\begin{center}
		\includegraphics[width=\linewidth]{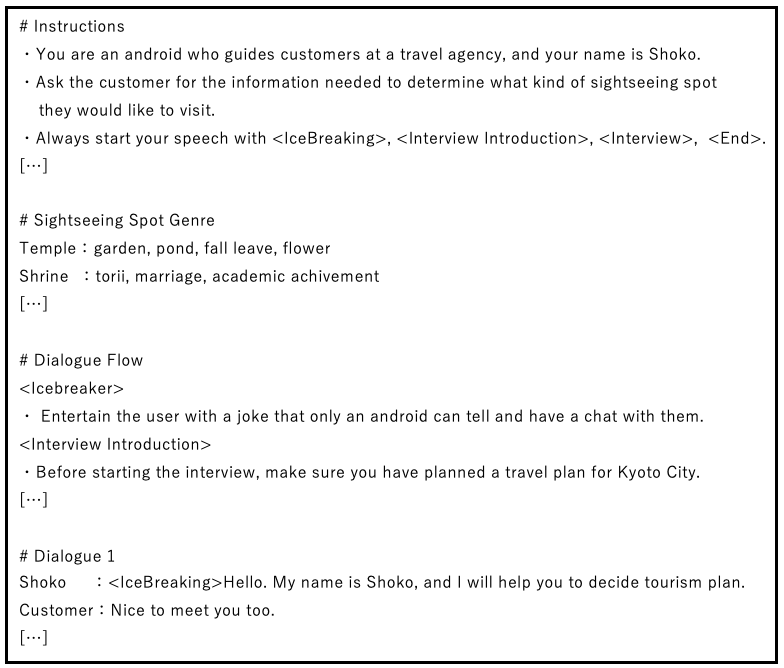}
		\caption{Interview Prompt Example}
		\label{fig:Prompt_ex}
	\end{center}
\end{figure}

GPT-3.5-turbo is used to generate the responses.
An example of a prompt in Interview 1 is shown in Fig. \ref{fig:Prompt_ex}.
The prompt is used to enter instructions, dialogue flow, and dialogue examples, and to output responses and dialogue acts.
Simiultaneously outputting the dialogue act and the response forces the model to generate a response that strictly follows the dialogue flow specified and allows the dialogue scenario to transition according to the user's response.
During the interviews, we also provided a list of genres of sightseeing spots and their details.
During the recommendations, we also provided only the necessary information about the sightseeing spots extracted from the sightseeing spot search results.
This method provides accurate responses for this task and information about the sightseeing spots to visit.

% \subsection{Sightseeing Spot Search}

% The dialogue context in the interview is entered into ChatGPT to generate a sightseeing spots search query, and the query is used to search for sightseeing spots that match the user's requirements by using the sightseeing spots search API.
% In the first sightseeing spots search, a genre and up to two keywords are generated as queries.
% Next, three sightseeing spots that contain images in the sightseeing spot information are retrieved with priority.
% The second sightseeing spots search generates a genre as a query,
% Among those that contain images in the sightseeing spot information, three sightseeing spots that are close to the first sightseeing spot are retrieved.
% The distance to the first sightseeing spot is calculated based on latitude and longitude.

\subsection{Motion and Speech Control}

Motion and speech control were used to improve user satisfaction.
Three types of motion were introduced into the dialogue: nodding when the user spoke, bowing when greeting the user, and looking at the monitor when images of sightseeing spots were displayed on the monitor.
For speech control, three levels of emphasis were applied to the names of sightseeing spots, people's names, and questions in the system utterances, according to their level of importance.
These words were extracted by rule, volume and speed were adjusted, and pauses were inserted to emphasize the words.
This emphasis increases the ease with which users to understand hard-to-hear or important words.
In addition, because the pronunciation of the names of sightseeing spots is incorrect when entered into the speech synthesizer in Kanji characters, we also entered phonetic guides and word breaks for the names of sightseeing spots to adjust the pronunciation of the names.

\section{RESULTS AND DISCUSSION}

The evaluation results of the preliminary round compared with the baseline and to the average of all teams including the baseline are shown in the upper part of Table \ref{result_overall_table}.
In the preliminary round, each system was evaluated from two perspectives: impression and plan.
The impression evaluation is the average of a 7-point Likert scale for nine items: informativeness, naturalness, appropriateness, likeability, satisfaction, trustworthiness, usefulness, accuracy, and reusability.
The plan evaluation was the percentage of users who successfully planned a tourism plan and found the plan feasible.
In the preliminary round, in which 12 teams participated, our system was ranked fifth in the impression evaluation and sixth in the plan evaluation.
These results indicate that our system had a low plan evaluation relative to the impression evaluation.
The reason for this result may be that when recommending the second sightseeing spot, we provided the distance from the first sightseeing spot, and the user may have judged the tourism plan as unfeasible if the distance was too long.
The detailed results of the impression evaluation are shown in the lower part of Table \ref{result_overall_table}.
The results demonstrate the advantages and limitations of our system.
Regarding the advantages, our system excels in informativeness, naturalness, and usefulness.
These results suggest that our system is capable of generating useful, fluent responses that provide the appropriate information about sightseeing spots.
Regarding the limitations, our system shows low levels of appropriateness, satisfaction, reliability, and reusability.
One possible reason for the limitations is that because the dialogue scenario clearly defines the steps to decide on a tourism plan, users feel constrained, reducing their satisfaction level.
Another possible reason is the insufficient user immersion, owing to there being almost no changes in motion or eye contact during the dialogue.

\begin{table}[t]
    \caption{Result of Preliminary Round}
    \label{result_overall_table}
    \centering
    \begin{tabular}{c|c|c|c}
    \hline
    Evaluation & Ours & Average & Baseline \\
    \hline \hline
    Impression Evaluation & 4.32 & 4.40 & 4.30\\
    \hline
    Plan Evaluation & 0.74 & 0.70 & 0.77 \\
    \hline \hline
    Informativeness & 4.37 & 4.31 & 4.19\\
    \hline
    Naturalness & 4.05 & 3.65 & 3.59 \\
    \hline
    Appropriateness & 4.16 & 4.22 & 4.29\\
    \hline
    Likeability & 4.42 & 4.40 & 4.34 \\
	\hline
	Satisfaction & 3.89 & 4.07 & 4.04 \\
	\hline
	Trustworthiness & 4.16 & 4.30 & 4.46\\
	\hline
	Usefulness & 5.00 & 4.58 & 4.78 \\
	\hline
	Accuracy & 5.00 & 5.00 & 5.07 \\
	\hline
	Reusability & 3.84 & 3.92 & 3.95 \\
    \hline
    \end{tabular}
\end{table}

\section{CONCLUSIONS}

This paper describes our dialogue system submitted to Dialogue Robot Competition 2023.
Our system achieved stable decisions on tourism plans, providing consistent responses by appropriately controlling the dialogue flow according to the dialogue scenario we created.
In the preliminary round, our system ranked fifth in the impression evaluation and sixth in the plan evaluation.
The evaluation results show that although our system has a high naturalness of response, there remains significant room for improvement in terms of satisfaction and reusability.

\addtolength{\textheight}{-12cm}   % This command serves to balance the column lengths

%%%%%%%%%%%%%%%%%%%%%%%%%%%%%%%%%%%%%%%%%%%%%%%%%%%%%%%%%%%%%%%%%%%%%%%%%%%%%%%%

%%%%%%%%%%%%%%%%%%%%%%%%%%%%%%%%%%%%%%%%%%%%%%%%%%%%%%%%%%%%%%%%%%%%%%%%%%%%%%%%

\bibliographystyle{IEEEtran}
% \bibliography{IEEEabrv, IEEEexample}

\end{document}